\documentclass{article}

\usepackage{microtype}
\usepackage{graphicx}
\usepackage{subfigure}
\usepackage{booktabs} 
\usepackage{tablefootnote}

\usepackage{hyperref}

\usepackage[accepted]{icml2024}

\usepackage{amsmath}
\usepackage{amssymb}
\usepackage{mathtools}
\usepackage{amsthm}
\usepackage{times}
\usepackage{latexsym}
\usepackage{graphicx}
\usepackage{booktabs}
\usepackage{subcaption}
\usepackage{multirow}
\usepackage{pifont}

\newcommand{\cmark}{\ding{51}} 
\newcommand{\xmark}{\ding{56}} 

\usepackage[capitalize,noabbrev]{cleveref}

\theoremstyle{plain}

\theoremstyle{definition}

\theoremstyle{remark}

\usepackage[textsize=tiny]{todonotes}

\icmltitlerunning{M3Face: A Unified Multi-Modal Multilingual Framework for Human Face Generation and Editing}

\begin{document}

\twocolumn[

\icmltitle{M\textsuperscript{3}Face: A Unified Multi-Modal Multilingual Framework \\ for Human Face Generation and Editing}



\icmlsetsymbol{equal}{*}

\begin{icmlauthorlist}
\icmlauthor{Mohammadreza Mofayezi}{equal,*}
\icmlauthor{Reza Alipour}{equal,*}
\icmlauthor{Mohammad Ali Kakavand}{equal,*}
\icmlauthor{Ehsaneddin Asgari}{+}
\end{icmlauthorlist}

\icmlaffiliation{*}{Sharif University of Technology, Tehran, Iran.\\}
\icmlaffiliation{+}{Qatar Computing Research Institute, Doha, Qatar.\\}

\icmlcorrespondingauthor{}{easgari@hbku.edu.qa}

\icmlkeywords{Machine Learning, ICML, Face, Diffusion Models, Multi-Modal, Editing}

\vskip 0.3in
]



\printAffiliationsAndNotice{\icmlEqualContribution} 

\begin{abstract}
Human face generation and editing represent an essential task in the era of computer vision and the digital world. Recent studies have shown remarkable progress in multi-modal face generation and editing, for instance, using face segmentation to guide image generation. However, it may be challenging for some users to create these conditioning modalities manually. Thus, we introduce \textbf{M\textsuperscript{3}Face}, a unified multi-modal multilingual framework for controllable face generation and editing. This framework enables users to utilize only text input to generate controlling modalities automatically, for instance, semantic segmentation or facial landmarks, and subsequently generate face images. We conduct extensive qualitative and quantitative experiments to showcase our framework’s face generation and editing capabilities.
Additionally, we propose the \textbf{M\textsuperscript{3}CelebA Dataset}, a large-scale multi-modal and multilingual face dataset containing high-quality images, semantic segmentations, facial landmarks, and different captions for each image in multiple languages. 
The code and the dataset will be released upon publication.
\end{abstract}

\section{Introduction}

\begin{figure}[t]
    \centering
    \includegraphics[width=\columnwidth]{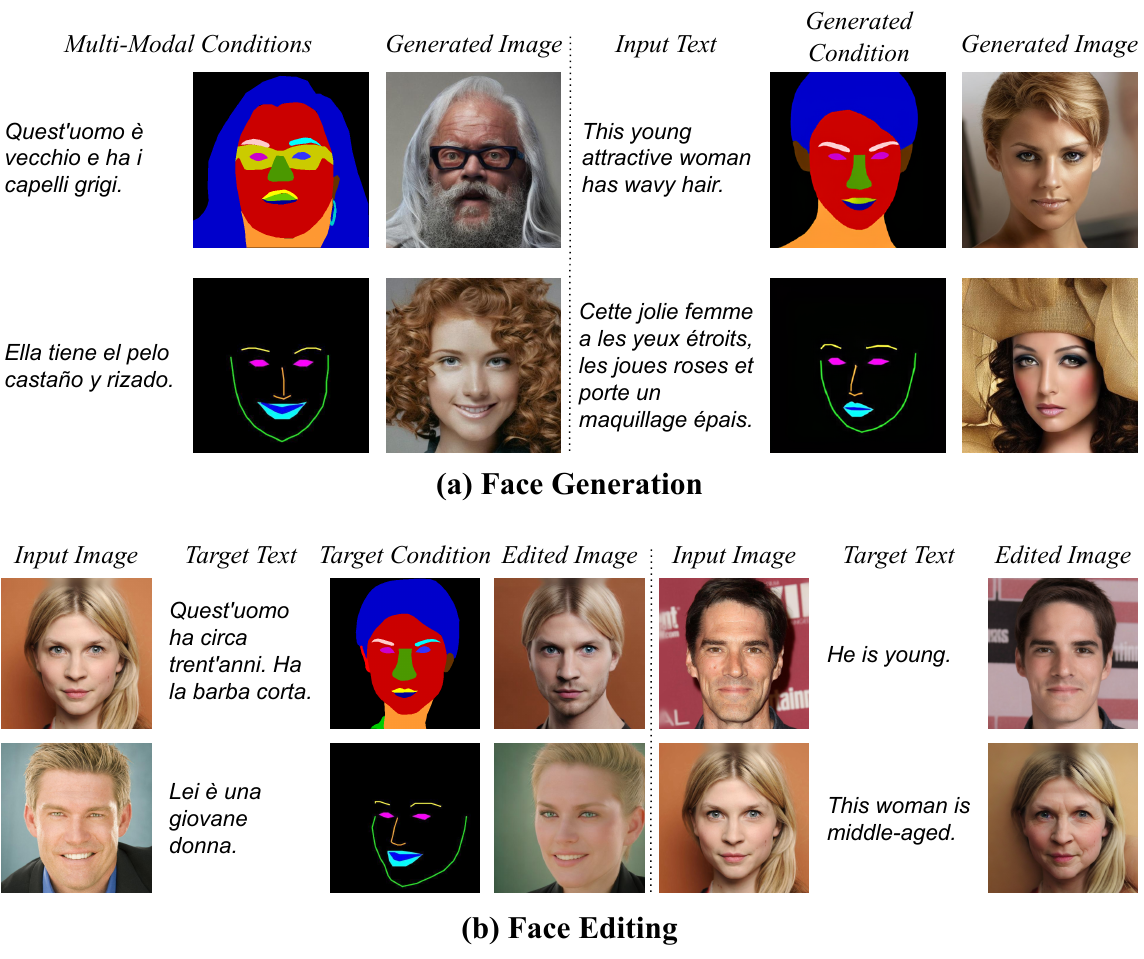}
    \caption{We introduce \textbf{M\textsuperscript{3}Face} for controllable multi-modal multilingual face generation and editing. \textbf{(a) Face Generation} can be done with both multi-modal conditions or a single text input. We generate face images consistent with the input text and other given modalities. \textbf{(b) Face Editing} can also be done with text, mask, landmarks, or a combination of them.}
    \label{fig:pull}
\end{figure}

The field of image generation and editing has witnessed remarkable progress, propelled by advancements in generative models \cite{karras2019style, xia2021tedigan, dhariwal2021diffusion, park2023learning} and large-scale datasets \cite{liu2015faceattributes, CelebAMask-HQ, xia2021tedigan}. These advancements are supposed to substantially contribute to the forthcoming generation of animations, movies, and video games \cite{guo2023animatediff, khachatryan2023text2video}. The creation of human characters, particularly characterized by their facial features, is one of the fundamental elements in the field of Generative AI. The innovations in this domain have given rise to sophisticated techniques capable of creating and modifying human facial images with unprecedented realism and versatility. Beyond their synthesis capability, these methods exhibit proficiency in handling diverse modalities, such as text \cite{reed2016generative, rombach2022high, chang2023muse, saharia2022photorealistic, jiang2021talk, kawar2023imagic, mokady2023null}, semantic segmentation masks \cite{zhang2023adding, wang2022semantic, rombach2022high}, facial landmarks \cite{zhang2023adding, zhang2020freenet, tang2021total}, or a combination of them \cite{xia2021tedigan, huang2023collaborative}.

Using multiple modalities to guide the generation of face images is a powerful approach. This allows users to precisely control and customize various facial features in the generated images. However, it poses some challenges; for instance, creating specific modalities manually, such as semantic segmentation, can be a complex task for most users. The intricate process of generating these conditions from scratch can be a potential barrier to the ease of use of multi-modal approaches.



In addition to the advancements in multi-modal generation and editing, the exploration of multilingual features has become more prominent \cite{ye2023altdiffusion, mclip, saxon2023multilingual}. Acknowledging the global diversity of languages and cultures, researchers are now incorporating linguistic elements into the generation and editing processes. This multilingual perspective transcends language barriers and fosters a more inclusive and accessible approach to face image generation and editing.

In this paper, we introduce \textbf{M\textsuperscript{3}Face}, a unified multi-modal multilingual framework for controllable face generation and editing. While previous methods do well on face generation and editing with different modalities, users may not have the initial conditions in real-world scenarios. Our framework addresses these issues in two significant ways. Firstly, it simplifies the generation of necessary conditions for image generation, such as semantic segmentation, by a text prompt. The user can continue generating or editing these conditions interactively. The framework then incorporates this text input and generated conditions to generate face images utilizing our ControlNet \cite{zhang2023adding} model or edit face images with the Imagic \cite{kawar2023imagic} method, making it easier for users to control and generate the desired images. 
Furthermore, we propose the \textbf{M\textsuperscript{3}CelebA Dataset} built upon the original CelebA dataset \cite{liu2015faceattributes}. Our dataset contains over 150,000 images of faces (5 times larger than the Multi-Modal CelebA-HQ \cite{xia2021tedigan} dataset) and is designed to be used in various research fields. Secondly, \textit{M\textsuperscript{3}Face} is trained using multiple languages existing in \textit{M\textsuperscript{3}CelebA Dataset}, which makes it accessible to users all over the world.
To summarize, our contributions are as follows:
\begin{itemize}
    \item We introduce \textbf{M\textsuperscript{3}Face}, a unified multi-modal multilingual framework for controllable face generation and editing. Unlike other multi-modal methods, this method can generate necessary conditions by a text prompt, offering an option to use additional conditioning modalities without requiring them. Additionally, this framework supports multiple languages.
    \item We propose the \textbf{M\textsuperscript{3}CelebA Dataset}, a large-scale multi-modal and multilingual face dataset containing high-quality images, semantic segmentations, facial landmarks, and different multilingual captions for each image.
    \item We achieve state-of-the-art qualitative and quantitative results in face generation and editing using facial landmarks and semantic segmentation.
\end{itemize}

\section{Related Work}

\noindent \textbf{Face Generation.}~~
Various image generation approaches have been explored, employing diverse architectures to enhance the quality and controllability of image generation. 
GAN-based methods can generate images with text by leveraging their latent space. TediGAN \cite{xia2021tedigan} added controllability to the generation by utilizing different modalities, which led to more stable results. Stable Diffusion \cite{rombach2022high} achieved improved qualitative results with an easier learning process in image generation. Similar success has also been found in various related tasks, including face generation. Collaborative Diffusion \cite{huang2023collaborative} uses pre-trained uni-modal diffusion models to utilize semantic segmentation maps as guidance in image generation. GCDP \cite{park2023learning} uses a Gaussian-categorical diffusion process to learn the joint image-semantic distribution and generate image-semantic pairs with text.

\noindent \textbf{Face Editing.}~~
Many works have tried to enhance the quality and accuracy of face image editing as a complicated task. 
Talk-to-Edit \cite{jiang2021talk} uses a location-specific semantic field in the GAN latent space to find the edit distance in this space. ChatFace \cite{yue2023chatface} learns high-level semantic information about the images and utilizes a diffusion model to edit images in this space. \cite{mokady2023null} is based on Stable Diffusion and uses Null-Text optimization for editing real images. 
Some methods use semantic segmentation as a guidance in editing face images. Collaborative Diffusion \cite{huang2023collaborative} uses semantic segmentation in image editing alongside text input in two diffusion models and then aggregates the results. Trying to utilize the pose and spatial properties of a face image, some methods tried to use keypoints in face editing. FReeNet \cite{zhang2020freenet} uses an encode-decoder architecture to edit input images with targeted landmarks. C\textsuperscript{2}GAN \cite{tang2021total} uses two GAN models to apply facial landmarks to the images.
While these methods benefit from different modalities to achieve stability and quality, they often pose challenges to user experience due to the intricate process of creating inputs like segmentation maps. In contrast, our method utilizes these modalities in the generation process by first creating them based on the input text and then using them in the latter parts of the pipeline.

\section{M\textsuperscript{3}Face}

\begin{figure*}
    \centering
    \includegraphics[width=\textwidth]{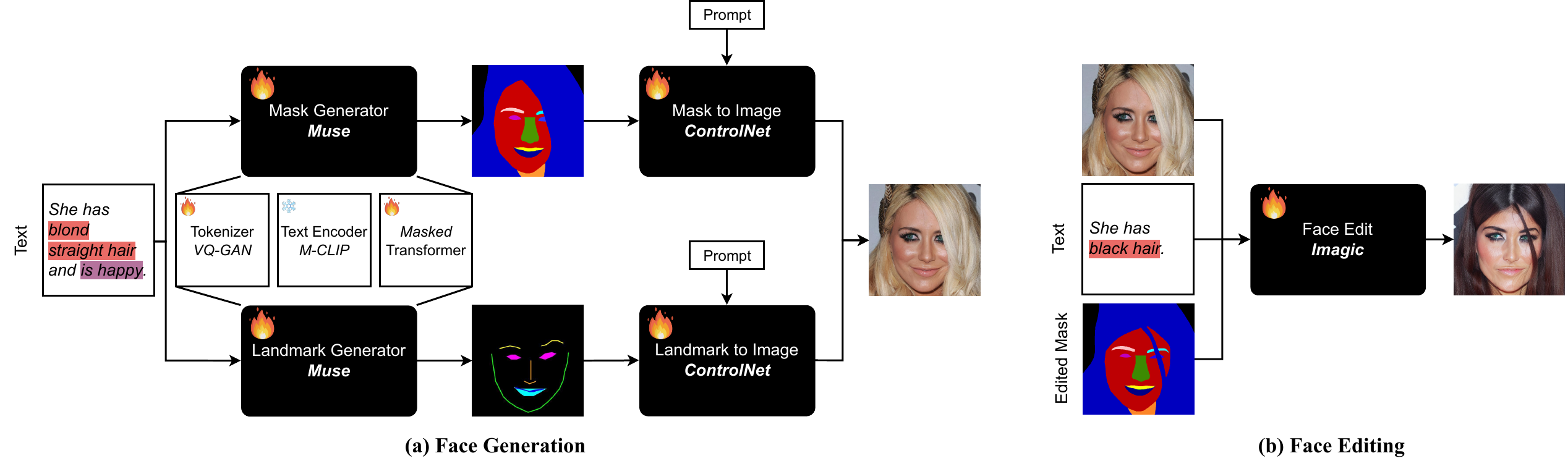}
    \caption{\textbf{M\textsuperscript{3}Face Framework.} For \textbf{(a) Face Generation}, we first generate facial landmarks or semantic segmentation masks with a given text input with our Muse \cite{chang2023muse} model. We then utilize our ControlNet \cite{zhang2023adding} model to generate face images from the intermediate results. For \textbf{(b) Face Editing}, we utilize the Imagic \cite{kawar2023imagic} method to manipulate images generated by our ControlNet. Face editing can be done with text, mask, landmarks, or a combination of them.}
    \label{fig:m3face-pipeline}
\end{figure*}

We introduce \textbf{M\textsuperscript{3}Face}, a unified framework for controllable multilingual human face generation and editing by immediately generating and editing facial landmarks and semantic segmentation. 
Figure \ref{fig:m3face-pipeline} shows an overview of the framework.

\subsection{Face Generation Pipeline}
For face generation, we first generate facial landmarks or semantic segmentation with a given text input with our masked transformer model inspired from \cite{chang2023muse}. 
We used a pre-trained Muse \cite{patil2024amused} model and adapted it to our usage. We employ a VQ-GAN \cite{Esser_2021_CVPR} for the task of tokenizing images. This involves mapping images into a sequence of tokens using a learned codebook. We fine-tuned this VQ-GAN model specifically for tokenizing segmentation masks and landmarks.
Our pipeline was designed to support multiple languages, and thus, we used the M-CLIP (LaBSE) \cite{mclip, feng2022languageagnostic} text encoder for the Muse architecture. We used this model instead of the T5 \cite{raffel2023exploring} model because the pre-trained Muse model used the CLIP-L14 \cite{radford2021learning} as a text encoder. So, we used this M-CLIP model to use the same pre-trained model.
Also, a U-VIT \cite{hoogeboom2023simple} architecture is used for the Muse base transformer model. Fine-tuning is performed on this model to adapt it to the specific requirements of our pipeline for generating masks and landmarks. After fine-tuning VQ-GAN and changing the text encoder, we fine-tuned this model to generate masks and landmarks.

After generating a mask or a landmark, we utilize a ControlNet \cite{zhang2023adding} architecture to generate face images from the intermediate results. We train two distinct ControlNet models: one for generating images based on landmark conditions and another for generating images based on mask conditions. These models are integrated into the M\textsuperscript{3}Face pipeline, and we can choose which method we want to choose based on our usage of the facial landmarks or the face segmentation.
Once a segmentation mask or landmark is generated using Muse, the corresponding ControlNet model will be used to create the final face images. 

\subsection{Face Editing Pipeline}
Face manipulation can be done with text, masks, landmarks, or a combination of them.
We first edit the facial landmarks or segmentation mask using the inpainting technique with the mask/landmark Muse model. Unlike general image editing, where details might be lost with inpainting or the identity of an image might be changed, masks and landmarks serve as fundamental structural elements of a face. Using inpainting to edit these elements will not have these issues because these images are not very highly detailed, and a portion of these structures usually can provide sufficient information for reconstructing the original structure. 

Then, we utilize the Imagic \cite{kawar2023imagic} method to manipulate images with our trained ControlNet models. 
We optimize the embedding of the editing prompt and then fine-tune the UNet model in the LDM part of our ControlNet model. We directly input the edited mask into the ControlNet model. Unlike Collaborative Diffusion \cite{huang2023collaborative}, which requires multiple steps to apply this method for different modalities, we perform the edit in a single step. This approach makes the multi-modal (or uni-modal) edit faster and results in a better outcome.

\section{M\textsuperscript{3}CelebA Dataset}
We propose the \textbf{M\textsuperscript{3}CelebA Dataset}, a large-scale multi-modal multilingual face dataset based on the original CelebA \cite{liu2015faceattributes} dataset. Our dataset contains more than 150K face images with semantic segmentation, facial landmarks, and multilingual captions.

\begin{figure}
    \centering
    \includegraphics[width=\columnwidth]{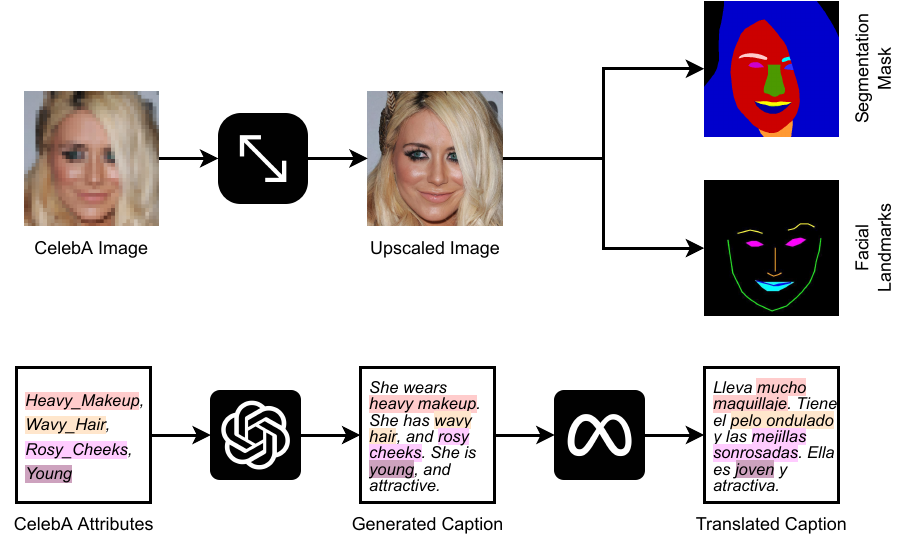}
    \caption{\textbf{Dataset Generation Pipeline.} We first align, crop, and upscale the original CelebA images. We then generate the facial landmarks and semantic segmentation for each image. For generating the captions, we use the 40 CelebA attributes and utilize the GPT3.5 model for generation and the SeamlessM4T model for translation.}
    \label{fig:dataset-generation-pipeline}
\end{figure}

In Figure \ref{fig:dataset-generation-pipeline}, you can see an overview of the dataset generation pipeline. We first align and crop the original CelebA images using the facial landmark annotations included in the original dataset and then upscale it to $512\times512$ with the Real-ESRGAN \cite{wang2021realesrgan} model. For facial landmarks, we utilize the dlib \cite{dlib09} library, and for semantic segmentation, we use the DML-CSR \cite{Zheng2022DecoupledML} model. For generating the captions, we use the 40 annotated attributes in the original CelebA dataset. We create three captions for each image utilizing a fine-tuned GPT3.5 \cite{brown2020language} model for generation and the SeamlessM4T \cite{barrault2023seamlessm4t} model for translation. 
Finally, we filter our dataset by both human and automatic evaluation and remove images with occlusions, blurriness, or extreme poses.
Some samples from the dataset are shown in Figure \ref{fig:dataset-overview}.
More details about the dataset are available in Appendix \ref{sec:appendix-dataset}.
A sample from the dataset is also available in the supplementary materials.

\begin{figure}[t]
    \centering
    \includegraphics[width=\columnwidth]{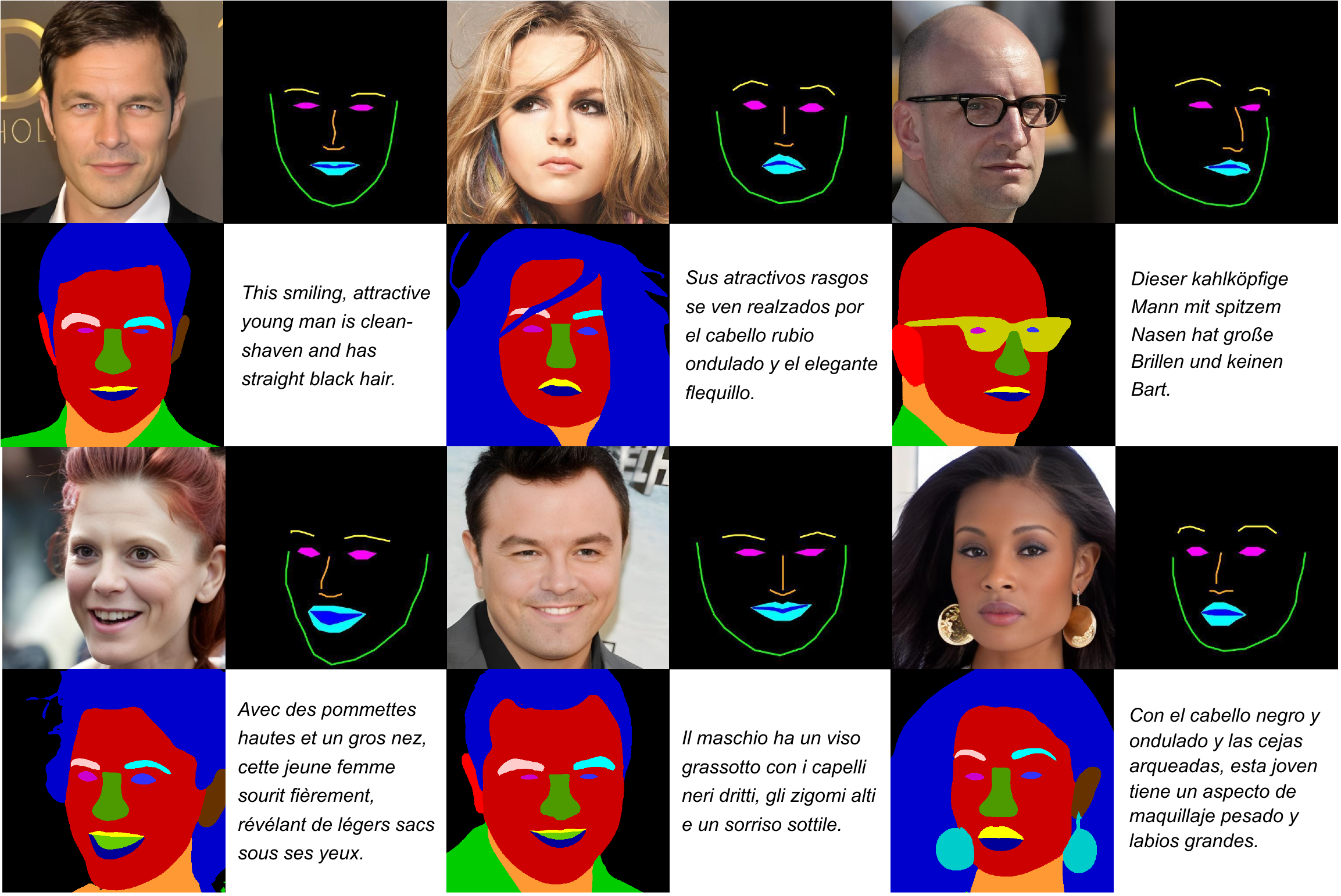}
    \caption{\textbf{M\textsuperscript{3}CelebA Dataset.} $512\times512$ images from the M\textsuperscript{3}CelebA dataset as well as the generated facial landmarks and semantic segmentation. Three multilingual captions are available for each image.}
    \label{fig:dataset-overview}
\end{figure}

\section{Experiments}

\begin{figure*}
    \centering
    \includegraphics[width=\textwidth]{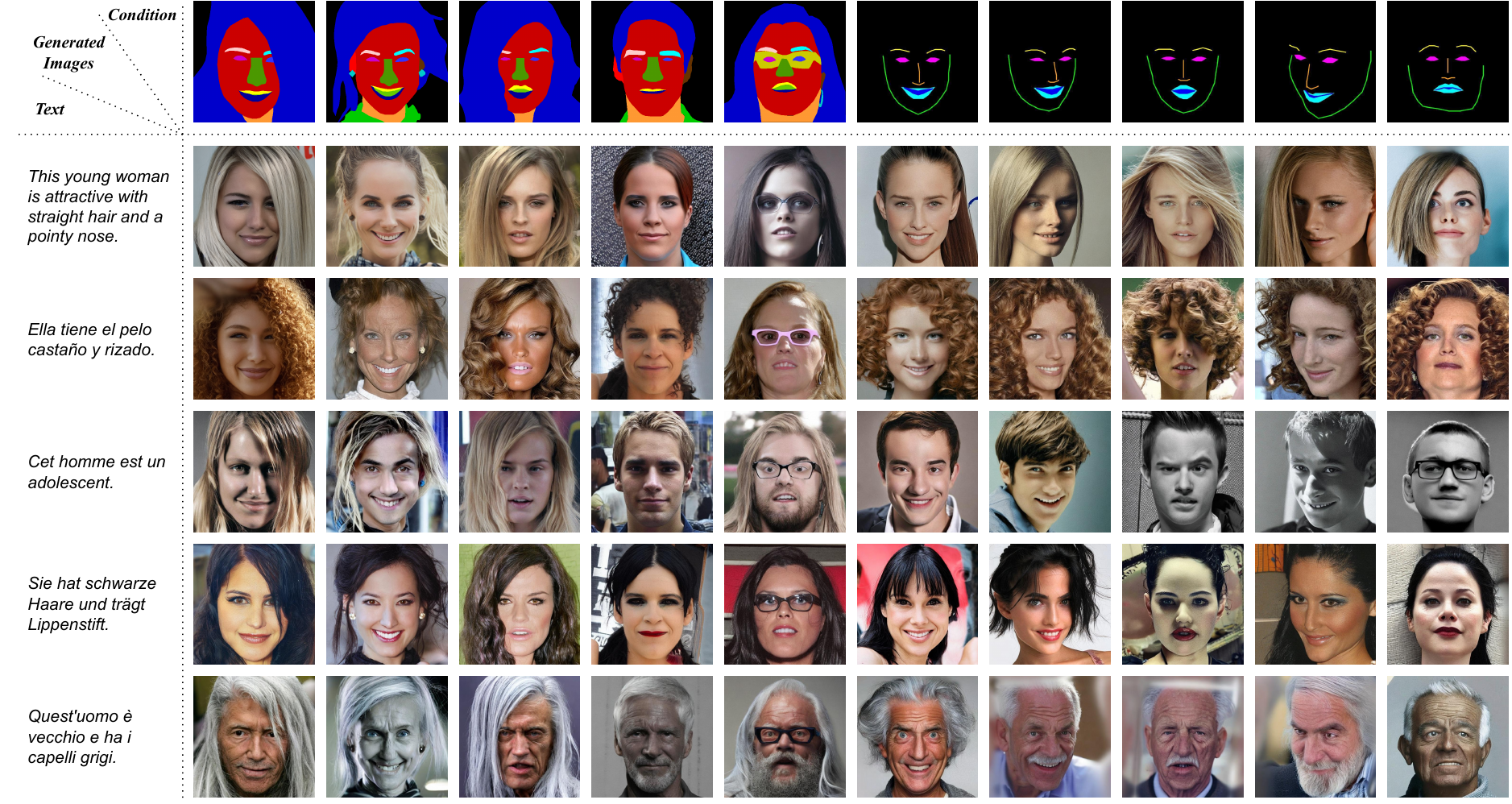}
    \caption{\textbf{Face Generation Results.} Our method generates realistic images based on the input prompt and the conditioning modality. We can generate faces consistent with semantic segmentation and facial landmarks. It also captures difficult attributes in the input prompt or the segmentation, such as glasses, hair color and style, and different face directions.}
    \label{fig:face-generation-result}
\end{figure*}

\subsection{Experimental Setup}
\noindent \textbf{Data Preparation.}~~
%
We used a combination of the CelebA-HQ and M\textsuperscript{3}CelebA datasets for our training. We made some changes to these datasets’ face segmentation colors. Specifically, we made the left and right colors symmetric; for instance, we used the same color for both eyebrows to use flip data augmentation later. For training our Muse model, we used random flip for data augmentation. Also, during training ControlNet, we used multiple data augmentation techniques, including random flips, changing color, brightness, and contrast of condition images. We randomly changed some colors in condition images for training the ControlNet model because the Muse model could generate invalid colors. So, the ControlNet model will still work well in our pipeline.
We evaluate all methods on the test set of our dataset.

\noindent \textbf{Training Details.}~~
For training our Muse model, we used an open-source implementation called aMUSEd \cite{patil2024amused} and its pre-trained models. Firstly, we fine-tuned the VQ-GAN on three types of images: face segmentation, facial landmarks, and face portrait. 
Then, we fine-tuned the base $256\times256$ resolution model for face segmentation, facial landmarks, and face portrait generation. For this purpose, we added instructions for each task to the images’ captions. After completing 15000 training steps on these tasks, we performed fine-tuning for 5000 additional steps on two individual models: one for landmark generation and the other for segmentation generation. The percentage of masked latent tokens was sampled from a cosine masking schedule for training base mode. 
\label{sec:training}
For training ControlNet, we used a variant of Stable Diffusion that supports multiple languages \cite{ye2023altdiffusion}. 
We trained the ControlNet model for ten epochs with a total batch size of 4 and gradient accumulation of 16, with a learning rate of $5\times10^{-5}$. We trained two individual models, one by conditioning on facial landmarks and another by conditioning on segmentation masks. To make the model more accurate on condition images, we used 25 percent of training samples without a caption and only with a simple prompt for generating a high-quality portrait of a face.

\noindent \textbf{Face Editing.}~~
For face editing with the Imagic \cite{kawar2023imagic} method, we use Adam \cite{kingma2014adam} optimizer with a learning rate of $10^{-3}$ for optimizing the text embedding and $5\times10^{-5}$ for fine-tuning the UNet model.
We optimize the embedding for 500 and fine-tune the model for 1000 steps.

\begin{figure*}[t]
    \centering
    \includegraphics[width=0.8\textwidth]{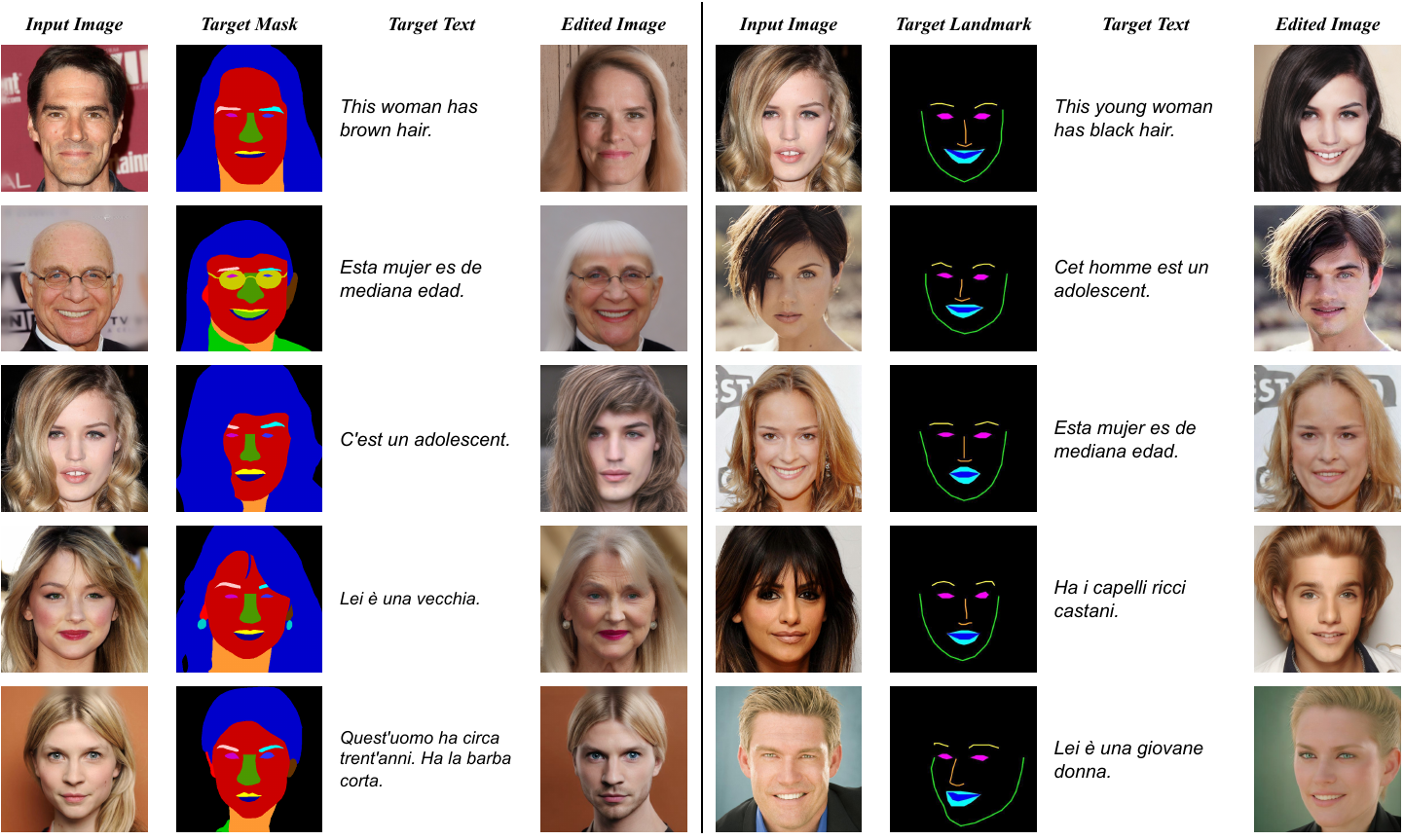}
    \caption{\textbf{Face Editing Results.} We can edit face images with both semantic segmentation and facial landmarks. The results are consistent with the conditioning and the text prompt. We achieve better results in details such as hair color and style.
}
    \label{fig:face-editing-result}
\end{figure*}

\subsection{Baselines}
We compare our proposed framework with state-of-the-art approaches for text-guided and multi-modal face generation and editing.

\textbf{Collaborative Diffusion} \cite{huang2023collaborative} is an LDM-based multi-modal face generation and editing method. They introduce a dynamic diffuser to combine different uni-modal pre-trained diffusion models. Similar to us, they utilize the Imagic method for multi-modal face editing.

\textbf{TediGAN} \cite{xia2021tedigan} is a StyleGAN-based face generation method. They utilize StyleGAN’s latent space to project the embedding of different modalities into it. The performs style mixing to achieve multi-modal face generation. 

\textbf{GCDP} \cite{park2023learning} is an Imagen-based text-guided image-mask pair generation method. 
They propose a Gaussian-categorical diffusion process that simultaneously generates images and corresponding semantic segmentations.

\textbf{Stable Diffusion} \cite{rombach2022high} is a latent text-to-image diffusion model. We compare face generation results with Stable-Diffusion-v1-5, which was initialized with the weights of the Stable-Diffusion-v1-2 checkpoint and subsequently fine-tuned with more steps.

\textbf{Talk-to-Edit} \cite{jiang2021talk} performs interactive face editing via dialog. They model a location-specific semantic field. However, it is limited to several predefined attributes.
Similar to this work, ChatFace \cite{yue2023chatface} conducts text-driven face editing in the semantic latent space of a diffusion model. This work has no public implementation; thus we do not include them in the comparisons.

\textbf{Null-Text Inversion} \cite{mokady2023null} is a Stable Diffusion-based method for editing real images by using Null-text inversion, i.e., modifying the unconditional text embedding rather than the input text embedding.

\subsection{Evaluation Metrics}
\noindent \textbf{FID.}~~
The quality of generated images is evaluated through Frechet Inception Distance, which measures the feature representation's distance between generated images and real images. Lower FID implies better sample quality.

\noindent \textbf{CLIP Score.}~~
The CLIP Score is the cosine similarity between the normalized image and text embeddings. A higher score usually indicates higher consistency between the generated image and the text prompt. We utilized the MCLIP-XLM model \cite{mclip, conneau-etal-2020-unsupervised} to assess the scores, as we needed to compare models with captions in diverse languages.


\noindent \textbf{Directional CLIP Similarity.}~~
It measures the consistency of the change between the two images (in CLIP \cite{radford2021learning} space) with the change between the two image captions. This metric is used for evaluating image editing, and the higher the directional CLIP similarity, the better it is.

\noindent \textbf{Segmentation and Landmark Consistency.}~~
Segmentation Consistency is the pixel-wise accuracy against the ground-truth segmentation. Landmark Consistency is the distance between ground-truth landmarks and the generated face landmarks.

\begin{figure*}[t]
    \centering
    \includegraphics[width=\textwidth]{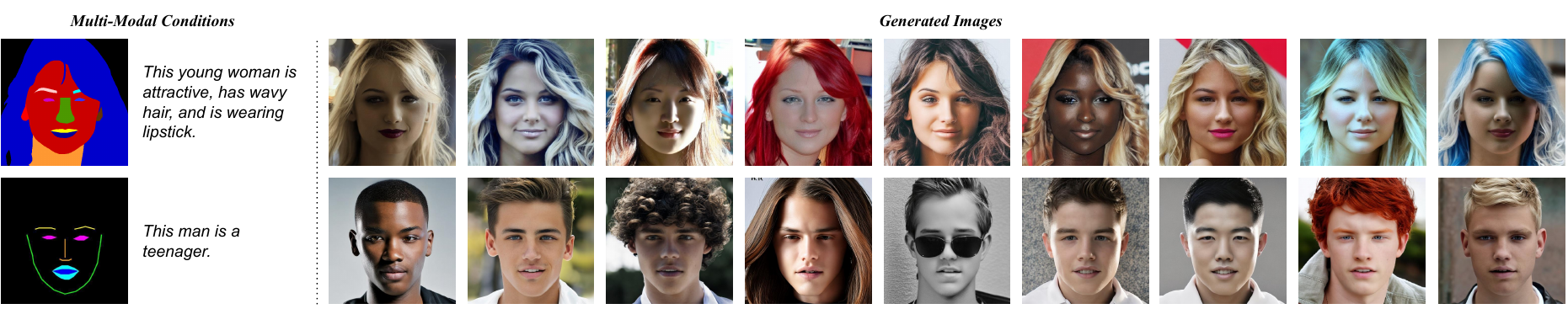}
    \caption{\textbf{Diversity of Face Generation.} Diverse face images can be generated from a single text input and a conditioning modality. We can see a diverse set of features such as hair colors, styles, skin tones, glasses, and hats.
}
    \label{fig:face-generation-diversity}
\end{figure*}

\begin{figure*}[t]
    \centering
    \includegraphics[width=\textwidth]{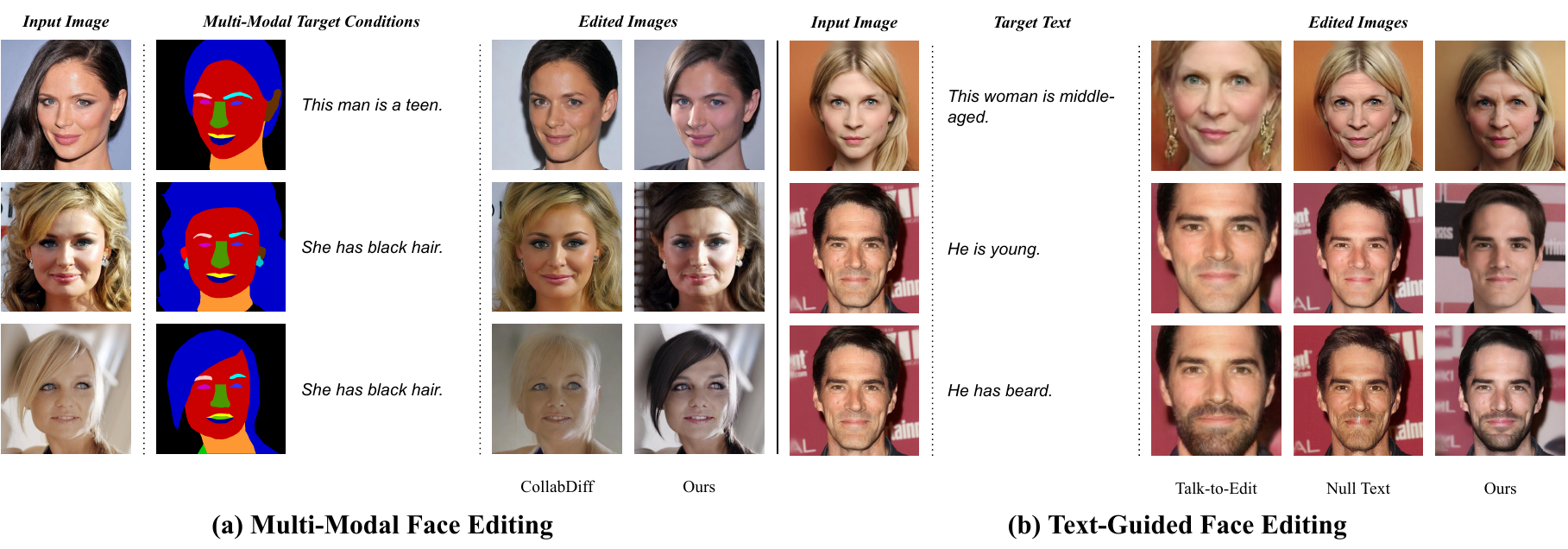}
    \caption{\textbf{Qualitative Comparison of Face Editing.} In \textbf{(a) Multi-Modal Face Editing}, while Collaborative Diffusion \cite{huang2023collaborative} fails to follow both the target prompt and segmentation mask in examples that require editing details such as the hair color or style, our method manipulates faces accurately and maintains the original face identity.
    In \textbf{(b) Text-Guided Face Editing}, Talk-to-Edit \cite{jiang2021talk} fails to apply the edit on most of the in-the-wild images and performs poorly on them. Null-Text Inversion \cite{mokady2023null} maintains the original face identity very well but is too sensitive to the edit prompt and sometimes fails to edit attributes accurately.
}
    \label{fig:face-editing-comparison}
\end{figure*}

\noindent \textbf{Human Evaluation.}~~
We conducted a user study to assess our framework's performance. For face generation, 200 multi-modal conditions were randomly selected from the M\textsuperscript{3}CelebA dataset's test split, and images were generated based on these conditions. Evaluators chose the best image based on photo realism and consistency with the conditions. 
For face editing, evaluators assessed identity preservation and the same criteria as in the face generation study using input and edited images along with edited conditions.

\subsection{Comparison with Baselines}
\begin{figure}[t]
    \centering
    \includegraphics[width=1\columnwidth]{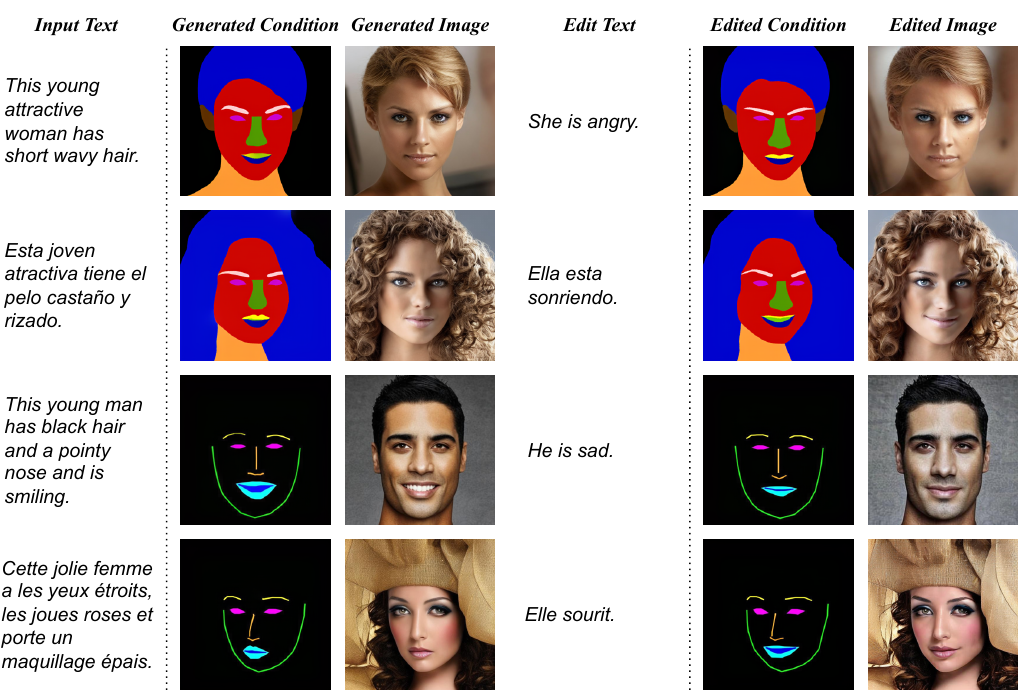}
    \caption{\textbf{Text-Guided Face Generation and Editing Results.} We can generate and edit face images with a single text prompt. We first generate semantic segmentation or facial landmarks according to the given text prompt. Then, we generate face images from these intermediate conditions. 
}
    \label{fig:face-generation-pipeline-result}
\end{figure}

\begin{figure}[!h]
    \centering
    \includegraphics[width=\columnwidth]{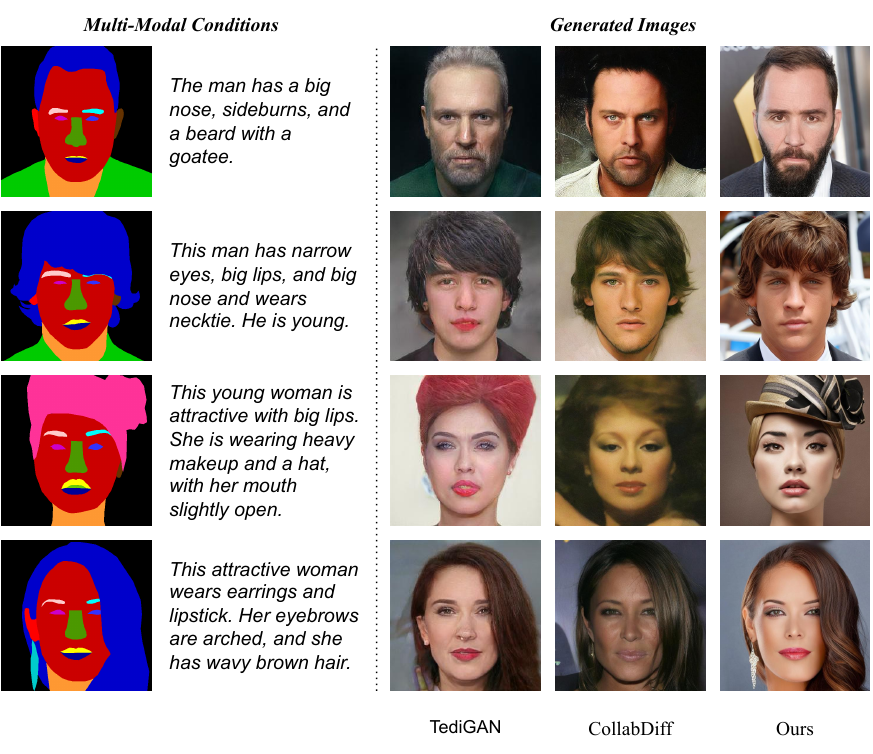}
    \caption{\textbf{Qualitative Comparison of Face Generation.} TediGAN \cite{xia2021tedigan} fails to follow both the target prompt and segmentation mask in most examples because of its two-step generation method. On the other hand, Collaborative Diffusion \cite{huang2023collaborative} often cannot generate details such as earrings, make-up, or a detailed beard.
}
    \label{fig:face-generation-comparison}
\end{figure}

\noindent \textbf{Qualitative Comparison.}~~
In this section, we show our face generation and editing results with both multi-modal and text-only conditions. In Figure \ref{fig:face-generation-result}, we provide multi-modal face generation results with semantic segmentation and facial landmarks conditions. We can generate diverse face images consistent with the given modalities as shown in Figure \ref{fig:face-generation-diversity}.
In Figure \ref{fig:face-generation-pipeline-result}, you can see the text-guided face generation and editing capabilities of our method. A semantic segmentation or facial landmarks condition will be generated for each input text prompt. Then, the face image can be generated with multi-modal conditions. We can see that both the generated conditions and the face images are consistent with the input prompt, and the model can generate or edit details such as emotions and face details.
For face generation, we compare our results with multi-modal methods \cite{huang2023collaborative, xia2021tedigan} in Figure \ref{fig:face-generation-comparison}. TediGAN \cite{xia2021tedigan} often fails to generate face images consistent with the given mask condition due to its two-step generation method. Collaborative Diffusion \cite{huang2023collaborative} also is not capable of generating details such as earrings, make-up, or specific beards like a goatee.

\begin{table}[t]
\centering
    \caption{\textbf{Quantitative Results of Face Generation.} Our method generates images with higher quality and more consistency with multi-modal conditions compared to TediGAN \cite{xia2021tedigan} and Collaborative Diffusion \cite{huang2023collaborative}. In generating faces from text input, our results surpass Stable Diffusion \cite{rombach2022high} and GCDP \cite{park2023learning}.}
    \setlength{\tabcolsep}{5pt}
    \begin{tabular}{lccccc}
        \toprule
        Method & FID $\downarrow$ & Text $\uparrow$ & Mask $\uparrow$ & Human $\uparrow$ \\ \midrule
        TediGAN & 58.49 & -\tablefootnote{TediGAN fine-tunes its model with the CLIP loss, and it eventually overfits on the input prompt. Thus, calculating the CLIP Score for this method is misleading.} & 0.90 & 19.02 \\ 
        CollabDiff & 38.20 & 24.80 & 0.90 & 23.64 \\
        Ours & \textbf{30.16} & \textbf{27.86} & \textbf{0.93} & \textbf{57.34} \\ \midrule
        SD-v1.5 & 67.29 & 26.82 & - & 8.22 \\
        GCDP & 53.45 & 27.15 & - & 10.32 \\ 
        Ours (Mask) & 39.14 & 27.28 & - & 23.93 \\
        Ours (Landmark) & \textbf{36.31} & \textbf{27.65} & - & \textbf{57.53} \\
        \bottomrule
    \end{tabular}
    \label{tab:face-generation-quantitative}
\end{table}

\begin{table}[t]
\centering
    \caption{\textbf{Quantitative Results of Face Editing.} We do better face editing than Collaborative Diffusion \cite{huang2023collaborative} in terms of segmentation consistency and directional CLIP similarity.}
    \setlength{\tabcolsep}{3.5pt}
    \begin{tabular}{lccc}
        \toprule
        Method & Mask/Landmark $\uparrow$ & Text $\uparrow$ & Human $\uparrow$ \\ \midrule
        CollabDiff & 0.91 & 0.04 & 19.80 \\
        Ours (Mask) & \textbf{0.94} & \textbf{0.09} & \textbf{80.20} \\
        Ours (Landmark) & 0.87 & 0.10 & - \\ \bottomrule
    \end{tabular}
    \label{tab:face-editing-quantitative}
\end{table}

In Figure \ref{fig:face-editing-result}, we show comprehensive multi-modal face editing results. We can see that the identity of faces is well-preserved and the generated faces are consistent with both the text and other conditions.
For face editing, we compare our results with both text-guided \cite{mokady2023null, jiang2021talk} and multi-modal \cite{huang2023collaborative} approaches. Figure \ref{fig:face-editing-comparison} shows the results. Collaborative Diffusion fails to follow the target mask and text prompt due to its two-step editing method. In Figure \ref{fig:face-editing-comparison} (b), we can see the text-guided editing results. While Null-Text Inversion \cite{mokady2023null} preserves the identity of the face very well, it struggles with editing specific attributes like beard and is sensitive to the target prompt. Talk-to-Edit \cite{jiang2021talk} has a problem with most in-the-wild images and cannot make edits to them. In successful attempts, it does not preserve the identity very well. 
Beyond these, we tried to compare with face editing methods using facial landmarks. FReeNet \cite{zhang2020freenet} seems to be very overfitted to specific poses and images, and the generated faces are either not changed or not reconstructed at all. 

\noindent \textbf{Quantitative Comparison.}~~
We compare our face generation results with both multi-modal and text-guided approaches in Table \ref{tab:face-generation-quantitative}. We outperform these methods in all quantitative metrics, showcasing the capabilities of our method.
We also compare multi-modal face editing with Collaborative Diffusion \cite{huang2023collaborative} in Table \ref{tab:face-editing-quantitative}.

\subsection{Ablation Study}
We study the effect of the $\alpha$ parameter used in the Imagic \cite{kawar2023imagic} method. Given an image and a target text prompt, we encode the target text and get the initial text embedding $e_{tgt}$, then optimize it to reconstruct the input image, obtaining $e_{opt}$. We then fine-tune the UNet model to improve fidelity to the input image while fixing $e_{opt}$. Finally, we interpolate between $e_{opt}$ and $e_{tgt}$ with an $\alpha$ coefficient to generate the edit result.
In Figure \ref{fig:ablation-alpha} (Appendix), the effect of this $\alpha$ coefficient and the number of UNet layers for fine-tuning is shown. The strength of the edit can be controlled with $\alpha$. Fine-tuning all of the UNet UpBlock parameters provides more consistent and controllable edits despite requiring more GPU memory compared to fine-tuning on fewer layers.




\section{Conclusion}
In this work, we propose a novel framework for multi-modal multilingual human face generation and editing called \textbf{M\textsuperscript{3}Face}. With this framework, users can generate face images and necessary conditions by a text prompt, offering the option to use additional conditioning modalities but not requiring them.
We also introduce the \textbf{M\textsuperscript{3}CelebA Dataset} which contains high-quality face images, semantic segmentations, facial landmarks, and multilingual captions for each image. 
We conduct extensive experiments to demonstrate our framework’s capabilities in face generation and editing with different modalities.
Our experiments show M\textsuperscript{3}Face's effectiveness, yet note a limitation with Muse architecture, which may produce invalid colors in segmentation and landmark generation. 
Another limitation regarding our framework is that the quality of generated images is highly affected by the Stable Diffusion backbone in the ControlNet model. 
Recent studies have introduced more robust SD models such as the Smooth Diffusion \cite{guo2023smooth}, which might improve face generation results and better preserve unedited content in face editing.
These investigations are left for future studies.


\section{Impact Statement}
The application of human face generation and editing holds considerable potential in creative domains. While it presents valuable opportunities, there is a potential for misuse, leading to the creation of harmful content or malicious editing of real human faces. Consequently, users are strongly urged to employ this technology responsibly and ethically.
\newpage
\bibliographystyle{icml2024}

\newpage
\appendix
\onecolumn
\section{Quantitative Results of Multilingual Face Generation}
We evaluate multilingual face generation and report the results in Table \ref{tab:multilingual-face-generation-quantitative}.
    
\begin{table}[!h]
\centering
    \caption{\textbf{Quantitative Results of Multilingual Face Generation.} Our method can generate high-quality face images with captions in different languages.}
    \begin{tabular}{lccc}
        \toprule
        Method & FID $\downarrow$ & Text $\uparrow$ & Mask $\uparrow$ \\ \midrule
        English & 30.16 & 27.86 & 0.93 \\
        Spanish & 30.51 & 27.31 & 0.93 \\
        French & 33.66 & 27.26 & 0.93 \\
        Italian & 32.93 & 27.35 & 0.93 \\
        German & 31.77 & 27.15 & 0.93 \\ \bottomrule
    \end{tabular}
    \label{tab:multilingual-face-generation-quantitative}
\end{table}

\section{Results of Ablation Study}
\begin{figure}[!h]
    \centering
    \includegraphics[width=0.8\columnwidth]{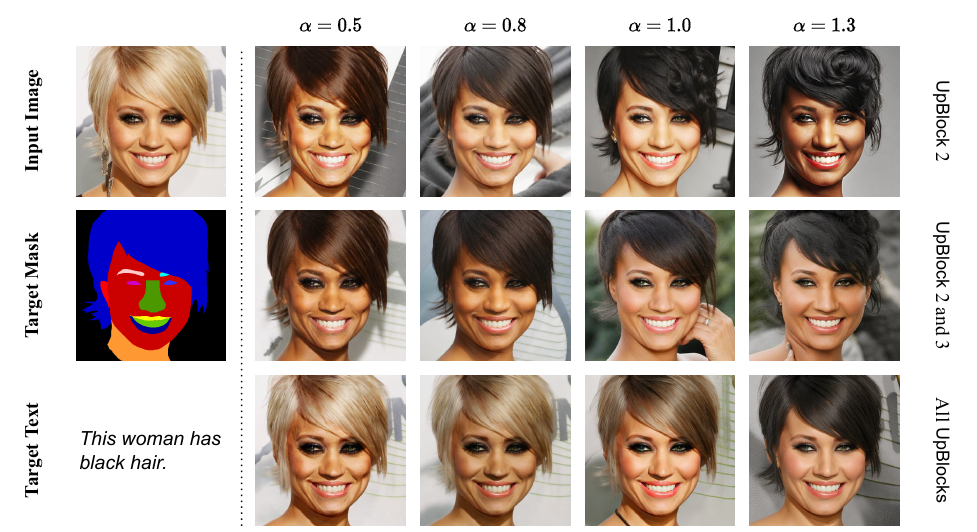}
    \caption{\textbf{Effect of Alpha and Number of Layers.} We can control the strength of textual editing with the alpha parameter. With more alpha values, we achieve higher face manipulation. While fine-tuning on fewer layers requires less GPU memory, we can have more consistent and controllable edits by fine-tuning all of the UNet UpBlock parameters.
}
    \label{fig:ablation-alpha}
\end{figure}

We show the results of our ablation study on the effect of different parameters on face editing in Figure \ref{fig:ablation-alpha}.
\section{Zero-Shot Face Generation}
\begin{figure}[!h]
    \centering
    \includegraphics[width=0.8\columnwidth]{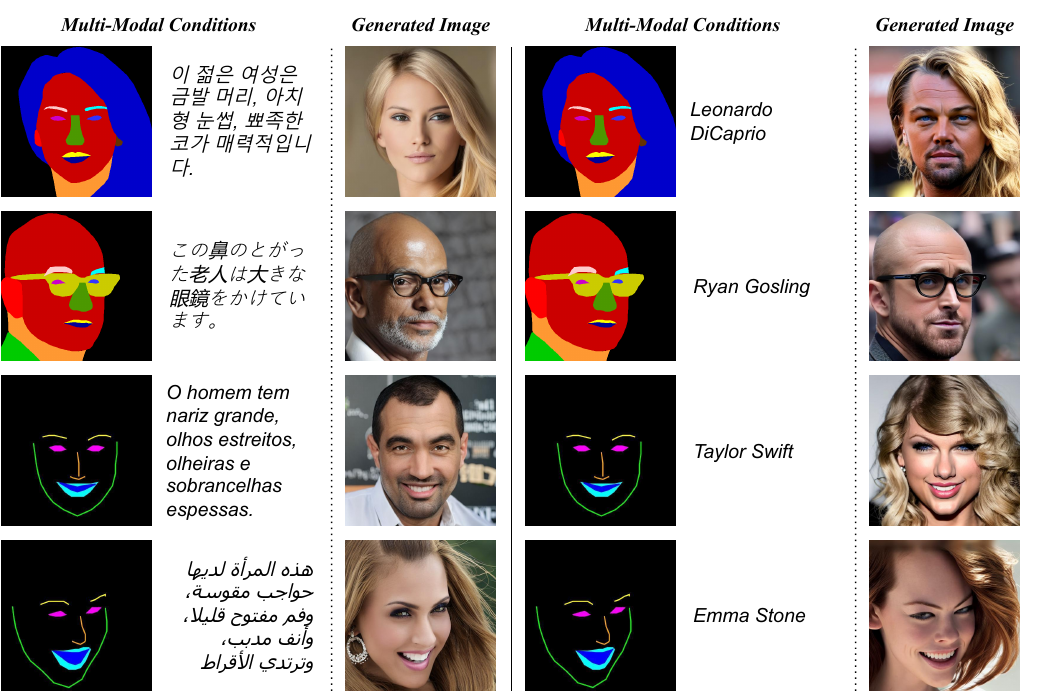}
    \caption{\textbf{Zero-Shot Generation Results.} Our pipeline can make zero-shot face generations with languages or prompts that are not in the training data.
}
    \label{fig:face-generation-zero-shot}
\end{figure}

In this section, we show the capabilities of the M\textsuperscript{3}Face pipeline in zero-shot face generation. We test captions in languages that are not in the training data, as well as specific human names. The results are shown in Figure \ref{fig:face-generation-zero-shot}.

\section{Details of the M\textsuperscript{3}CelebA Dataset}
\label{sec:appendix-dataset}
In this section, we explain our method for generating captions and the details and some statistics about our M\textsuperscript{3}CelebA dataset.

\begin{figure}[!h]
    \centering
    \includegraphics[width=\columnwidth]{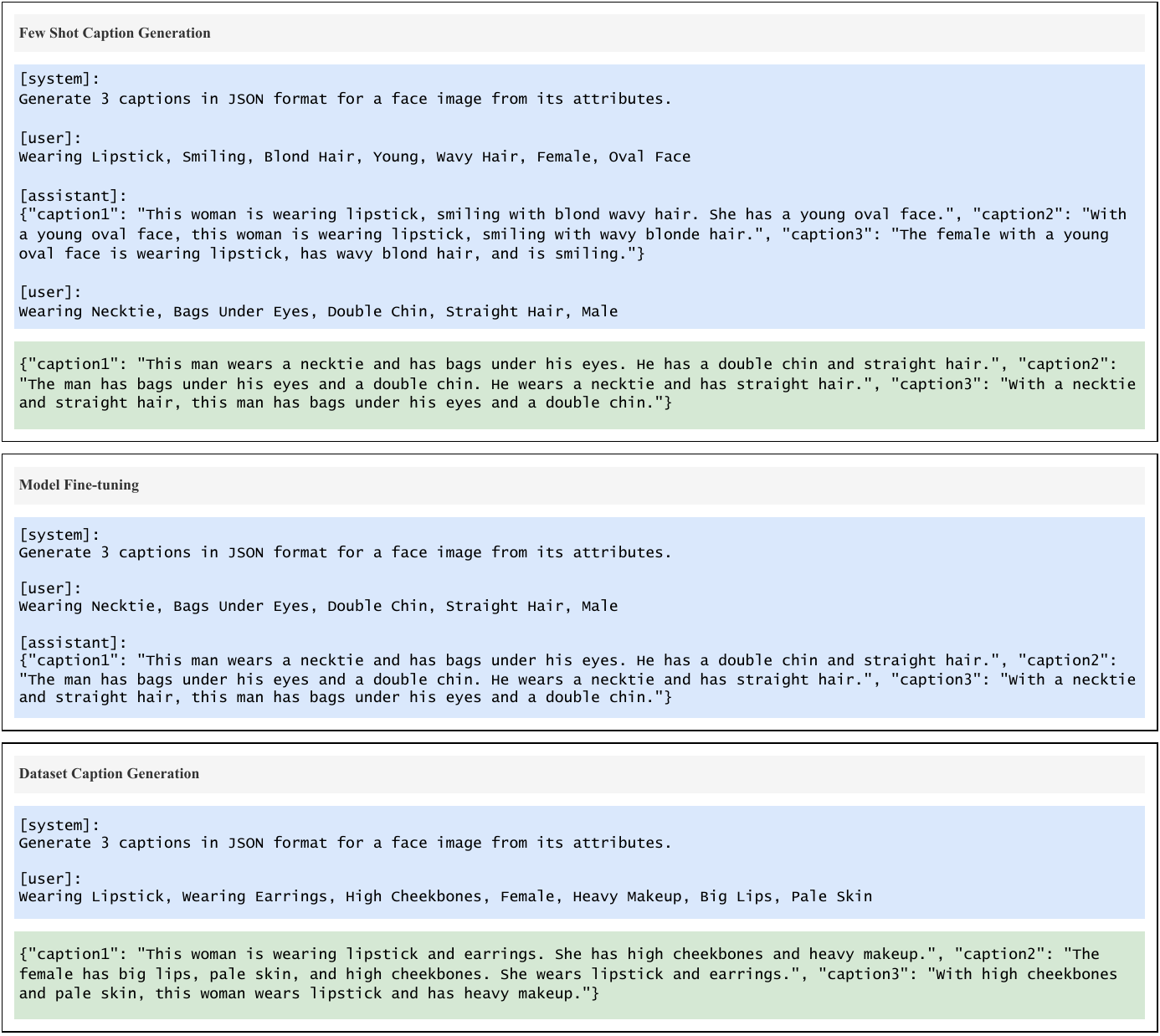}
    \caption{\textbf{Caption Generation Process.} The prompts used in the few-shot and fine-tuning processes are shown in this figure.
}
    \label{fig:caption-gpt}
\end{figure}

We used the few-shot technique with GPT-3.5 Turbo \cite{brown2020language} to generate captions for images based on their attributes. We then picked the top 1500 captions by human evaluation, which were subsequently used to fine-tune the GPT-3.5 Turbo. We fine-tuned the model for five epochs using the default training parameters. The fine-tuned model was then used to generate captions for all images. We show this process in Figure \ref{fig:caption-gpt}.

\begin{figure}[!h]
    \centering
    \includegraphics[width=\columnwidth]{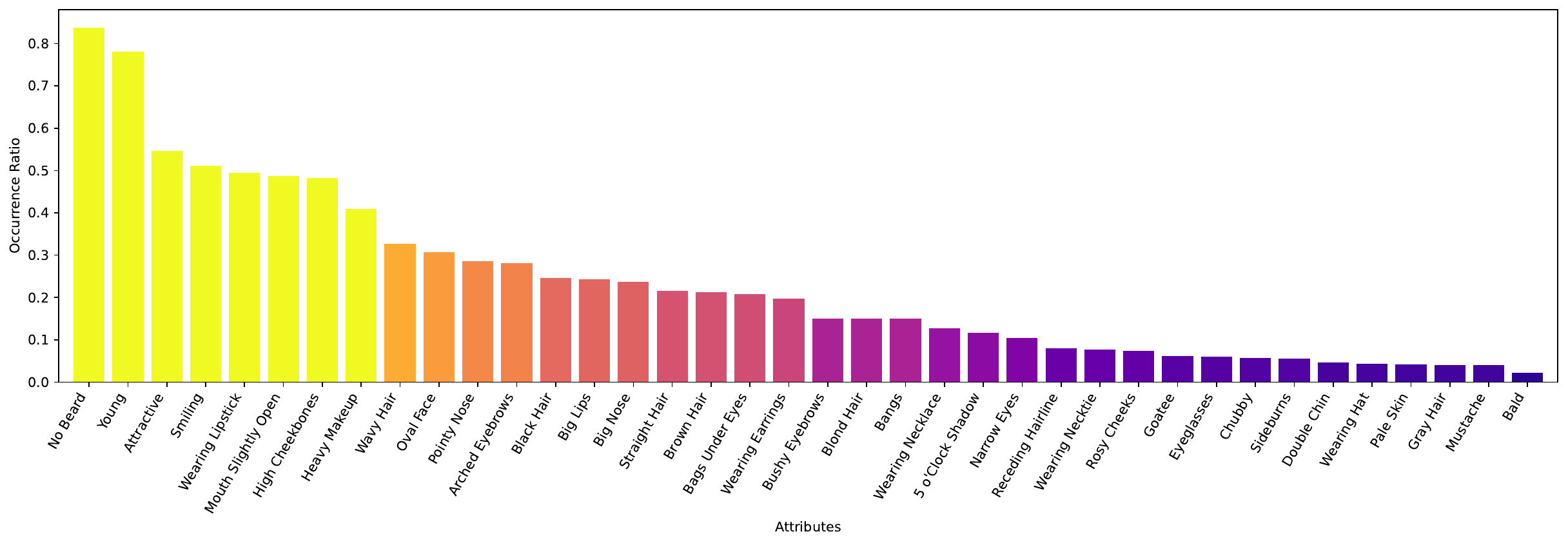}
    \caption{\textbf{Attribute Distribution in Dataset.} Occurrence ratio of different face attributes in M\textsuperscript{3}CelebA dataset.
}
    \label{fig:attribute-distribution}
\end{figure}

\begin{figure}[!h]
    \centering
    \includegraphics[width=0.8\columnwidth]{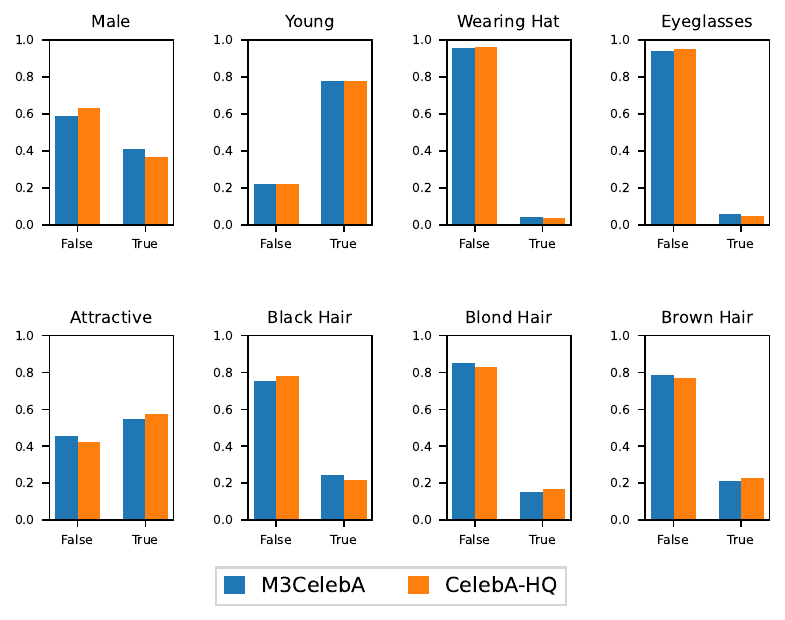}
    \caption{\textbf{Attribute Distribution Comparison.} We compare the occurrence ratio of different face attributes in CelebA-HQ and M\textsuperscript{3}CelebA datasets. Our dataset is more balanced in terms of Male/Female and Non-Attractive/Attractive attributes.
}
    \label{fig:attribute-comparison}
\end{figure}

\begin{figure}[!h]
    \centering
    \includegraphics[width=\columnwidth]{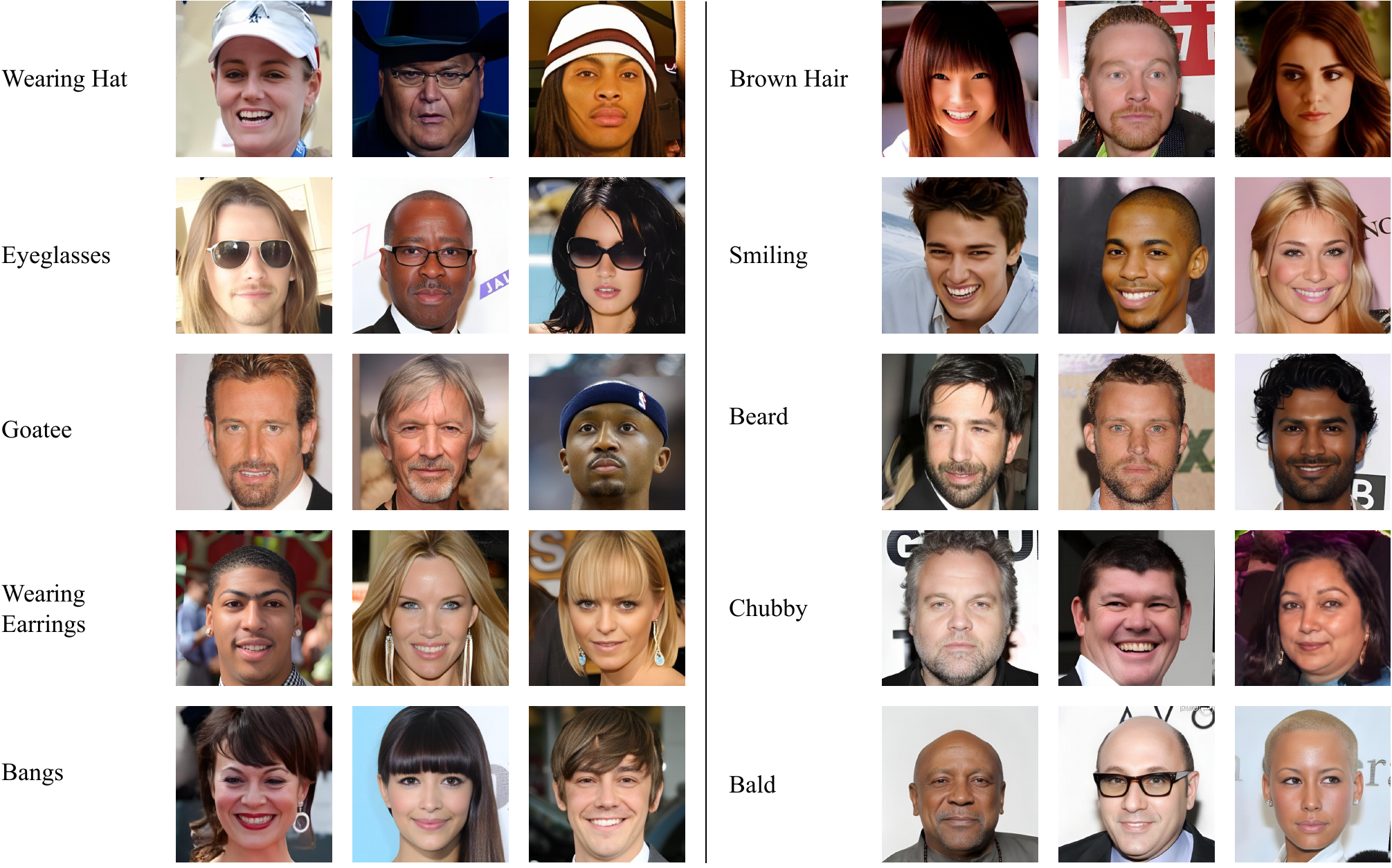}
    \caption{\textbf{Sample Images from Dataset.} Sample images with different face attributes from the dataset.
}
    \label{fig:dataset-attribute-samples}
\end{figure}

In Table \ref{tab:dataset-comparison}, we compare the available CelebA-based datasets.
The CLIP Score of our M\textsuperscript{3}CelebA captions is 3.1\% higher than the Multi-Modal CelebA-HQ \cite{xia2021tedigan} captions, showing that the quality of our generated captions is better.
We show the distribution of different attributes in the dataset in Figure \ref{fig:attribute-distribution}.
In Figure \ref{fig:attribute-comparison}, we compare the distribution of several important attributes to the CelebA-HQ dataset. We see improvements in several attributes such as Male/Female, Eyeglasses, and Black Hair. The number of face images with "Attractive" attributes has decreased, contributing to reducing bias in the dataset.
Finally, we show several example images from some of the attributes in Figure \ref{fig:dataset-attribute-samples}. 

\begin{table}[!h]
\centering
\caption{\textbf{Comparison of Different CelebA-Based Datasets.} We compare different characteristics of datasets created based on the CelebA \cite{liu2015faceattributes} dataset.}
\begin{tabular}{lccccc}
\toprule
Name & Size & Segmentation Map & Landmark & Caption & Multilingual \\
\midrule
CelebA & 202599 & \xmark & \xmark & \xmark & \xmark \\
CelebA-HQ & 30000 & \xmark & \xmark & \xmark & \xmark \\
CelebAMask-HQ & 30000 & \cmark & \cmark & \xmark & \xmark \\
CelebA-Dialog & 202599 & \xmark & \xmark & \cmark & \xmark \\
MM-CelebA-HQ & 30000 & \cmark & \cmark & \cmark & \xmark \\
M\textsuperscript{3}Face & 173314 & \cmark & \cmark & \cmark & \cmark \\
\bottomrule
\end{tabular}
\label{tab:dataset-comparison}
\end{table}

\end{document}